\documentclass[11pt,a4paper]{article}
\usepackage[hyperref]{acl2017}
\usepackage{times}
\usepackage{latexsym}

\usepackage{url}

\usepackage{color}
\usepackage{graphicx}
\usepackage{amsthm}
\usepackage{latexsym}
\usepackage{amsmath}
\usepackage{multirow}
\usepackage{amssymb}
\usepackage{bigstrut}
\usepackage{wrapfig}

\newcommand{\forCR}[1]{#1} 

\newcommand{\namecite}[1]{\citeauthor{#1}~\shortcite{#1}}

\newcommand{\tupleinf}{\textsc{TupleInf}}
\newcommand{\tableilp}{\textsc{TableILP}}
\newcommand{\ir}{\textsc{IR}}
\newcommand{\pmi}{\textsc{PMI}}

\newcommand\T{\rule{0pt}{2.6ex}}       
\newcommand\B{\rule[-1.2ex]{0pt}{0pt}} 
\aclfinalcopy 
\def\aclpaperid{3433} 
\title{Answering Complex Questions Using Open Information Extraction}

\author{Tushar Khot \\
 Allen Institute for AI \\
  {\tt tushark@allenai.org} \\\And
  Ashish Sabharwal \\
 Allen Institute for AI \\
  {\tt ashishs@allenai.org} \\\And
  Peter Clark \\
Allen Institute for AI \\
  {\tt peterc@allenai.org}
  }

\date{}

\begin{document}

\maketitle

\begin{abstract}
While there has been substantial progress in factoid question-answering (QA), answering complex questions remains challenging, typically requiring both a large body of knowledge and inference techniques. Open Information Extraction (Open IE) provides a way to generate semi-structured knowledge for QA, but to date such knowledge has only been used to answer simple questions with retrieval-based methods. We overcome this limitation by presenting a method for reasoning with Open IE knowledge, allowing more complex questions to be handled. Using a recently proposed support graph optimization framework for QA,
we develop a new inference model for Open IE, in particular one that can work effectively with multiple short facts, noise, and the relational structure of tuples. Our model significantly outperforms a state-of-the-art structured solver on complex questions of varying difficulty, while also removing the reliance on manually curated knowledge.

\end{abstract}

\section{Introduction}

Effective question answering (QA) systems have been a long-standing quest of AI research. Structured curated KBs have been used successfully for this task~\cite{sempre,parasempre}. However, these KBs are expensive to build and typically domain-specific. Automatically constructed open vocabulary \emph{(subject; predicate; object)} style tuples have broader coverage, but have only been used for simple questions where a single  
tuple suffices~\cite{oqa,taqa}.

Our goal in this work is to develop a QA system that can perform reasoning with Open IE~\cite{Banko2007OpenIE} tuples for complex multiple-choice questions that require tuples from multiple sentences. Such a system can answer complex questions in resource-poor domains where curated knowledge is unavailable. 
Elementary-level science exams is one such domain, requiring complex reasoning~\cite{aristo2015:challenge}.  
Due to the lack of a large-scale structured KB, state-of-the-art systems for this task either rely on shallow reasoning with large text corpora~\cite{aristo2016:combining,gaokaoIJCAI} or deeper, structured reasoning with a small amount of automatically acquired~\cite{aristo2015:mln} or manually curated~\cite{tableilp2016} knowledge.

Consider the following question from an Alaska state 4th grade science test:
\begin{quote}
\textit{Which object in our solar system reflects light and is a satellite that orbits around one planet? (A) Earth (B) Mercury (C) the Sun (D) the Moon}
\end{quote}

This question is challenging for QA systems 
because of its complex structure and the need for multi-fact reasoning. A natural way to answer it is by combining facts such as \emph{(Moon; is; in the solar system)}, \emph{(Moon; reflects; light)}, \emph{(Moon; is; satellite)}, and \emph{(Moon; orbits; around one planet)}. 

A candidate system for such reasoning, and which we draw inspiration from, is the \tableilp~system of \citet{tableilp2016}. \tableilp~treats QA as a search for an optimal subgraph that connects terms in the question and answer via rows in a set of curated tables, and solves the optimization problem using Integer Linear Programming (ILP). We similarly want to search for an optimal subgraph.
However, a large, automatically extracted tuple KB makes the reasoning context different on three fronts: (a) unlike reasoning with tables, chaining tuples is less important and reliable as join rules aren't available; (b) conjunctive evidence becomes paramount, as, unlike a long table row, a single tuple is less likely to cover the entire question; and (c) again, unlike table rows, tuples are noisy, making combining redundant evidence essential. Consequently, a table-knowledge centered inference model isn't the best fit for noisy tuples.

To address this challenge, we present a new ILP-based model of inference with tuples, implemented in a reasoner called \tupleinf. We demonstrate that \tupleinf~significantly outperforms \tableilp~by 11.8\% on a broad set of over 1,300 science questions, without requiring manually curated tables, using a substantially simpler ILP formulation, and generalizing well to higher grade levels. The gains persist even when both solvers are provided identical knowledge.
This demonstrates for the first time how Open IE based QA can be extended from simple lookup questions to an effective system for complex questions. 

\section{Related Work}

We discuss two classes of related work: retrieval-based web question-answering (simple reasoning with large scale KB) and science question-answering (complex reasoning with small KB). 

\paragraph{Web QA:}
There exist several systems for retrieval-based Web QA problems~\cite{watson,brill2002analysis}. While structured KBs such as Freebase have been used in many~\cite{sempre,parasempre,kwiatkowski13}, such approaches are limited by the coverage of the data. QA systems using semi-structured Open IE tuples~\cite{paralex,oqa,taqa} or automatically extracted web tables~\cite{tabCell,tableParse} have broader coverage but are limited to simple questions with a single query.

\paragraph{Science QA:}
Elementary-level science QA tasks require reasoning to handle complex questions.
 Markov Logic Networks~\cite{richardson2006markov} have been used to perform probabilistic reasoning over a small set of logical rules~\cite{aristo2015:mln}. Simple IR techniques have also been proposed for science tests~\cite{aristo2016:combining} and Gaokao tests (equivalent to the SAT exam in China)~\cite{gaokaoIJCAI}.  

The work most related to \tupleinf\ is the aforementioned \tableilp\ solver. This approach
focuses on building inference chains using manually defined join rules for a small set of curated tables. 
 While it can also use open vocabulary tuples (as we assess in our experiments),
 its efficacy is limited by the difficulty of defining reliable join rules for such tuples.
Further, each row in some complex curated tables
covers all relevant contextual information (e.g., each row of the \emph{adaptation} table contains (animal, adaptation, challenge, explanation)), whereas recovering such information requires combining multiple Open IE tuples.

\section{Tuple Inference Solver}

We first describe the tuples used by our solver. We define a tuple as \emph{(subject; predicate; objects)} with zero or more objects. We refer to the subject, predicate, and objects as the fields of the tuple.

\subsection{Tuple KB}
We use the text corpora (S) from ~\namecite{aristo2016:combining} to build our tuple KB. For each test set, we use the corresponding training questions $Q_\mathit{tr}$ to retrieve domain-relevant sentences from S. Specifically, for each multiple-choice question $(q,A) \in Q_\mathit{tr}$ and each choice $a \in A$, we use all non-stopword tokens in $q$ and $a$ as an ElasticSearch\footnote{https://www.elastic.co/products/elasticsearch} query against S. We take the top 200 hits, run Open IE v4,\footnote{http://knowitall.github.io/openie} and aggregate the resulting tuples over all $a \in A$ and over all questions in $Q_\mathit{tr}$ to create the tuple KB (T).\footnote{Available at \emph{http://anonymized}}

\subsection{Tuple Selection}
\label{sec:selection}
Given a multiple-choice question $qa$ with question text $q$ and answer choices A=$\{a_i\}$, we select the most relevant tuples from $T$ and $S$ as follows.

\textbf{Selecting from Tuple KB:} 
We use an inverted index to find the 1,000 tuples that have the most overlapping tokens with question tokens $tok(qa).$\footnote{All tokens are stemmed and stop-word filtered}. We also filter out any tuples that overlap only with $tok(q)$ as they do not support any answer. We compute the normalized TF-IDF score treating the question, $q$ as a query and each tuple, $t$ as a document\forCR{:}
\begin{align*}
&\textit{tf}(x, q)=1\; \textmd{if x} \in q ; \textit{idf}(x) = log(1 + N/n_x) \\
&\textit{tf-idf}(t, q)=\sum_{x \in t\cap q} idf(x)
\end{align*}
\forCR{where $N$ is the number of tuples in the KB and $n_x$ are the number of tuples containing $x$. We normalize the \textit{tf-idf} score by the number of tokens in $t$ and $q$.} We finally take the 50 top-scoring tuples $T_{qa}$~\footnote{\forCR{Available at http://allenai.org/data.html}}.

\textbf{On-the-fly tuples from text: }
To handle questions from new domains not covered by the training set, we extract additional tuples \emph{on the fly} from S (similar to \namecite{knowlhunting}). We perform the same ElasticSearch query described earlier for building T. We ignore sentences that cover none or all answer choices as they are not discriminative. We also ignore long sentences ($>$300 characters) and sentences with negation\footnote{containing not, 'nt, or except} as they tend to lead to noisy inference. We then run Open IE on these sentences and re-score the resulting tuples using the Jaccard score\footnote{$\mid tok(t) \cap tok(qa)\mid / \mid tok(t) \cup tok(qa) \mid$} 
due to the lossy nature of Open IE, and finally take the 50 top-scoring tuples $T'_{qa}$.

\subsection{Support Graph Search}

\begin{figure}[t]
\centering
\includegraphics[scale=0.24, trim=0px 10px 0px 0px, clip]{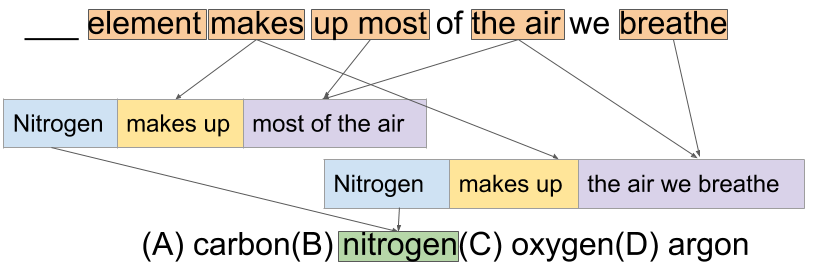}
\caption{An example support graph linking a question (top), two tuples from the KB (colored) and an answer option (nitrogen).}
\vspace{-4mm} 
\label{fig:suppGraph}
\end{figure}

Similar to \tableilp, we view the QA task as searching for a graph that best connects the terms in the question (qterms) with an answer choice via the knowledge; see Figure~\ref{fig:suppGraph} for a simple illustrative example. Unlike standard alignment models used for tasks such as  
Recognizing Textual Entailment (RTE)~\cite{dagan2010recognizing}, however, we must score alignments between a set $T_{qa} \cup T'_{qa}$ of structured tuples and a (potentially multi-sentence) multiple-choice question $qa$.

The qterms, answer choices, and tuples fields form the set of possible vertices, $\mathcal{V}$, of the support graph. Edges connecting qterms to tuple fields and tuple fields to answer choices form the set of possible edges, $\mathcal{E}$. The support graph, $G(V, E)$, is a subgraph of $\mathcal{G}(\mathcal{V}, \mathcal{E})$ where $V$ and $E$ denote ``active'' nodes and edges, resp. We define the desired behavior of an optimal support graph via an ILP model as follows\footnote{\forCR{c.f. Appendix~\ref{app_model} for more details}}.

\paragraph{Objective Function}\mbox{}\\
Similar to \tableilp, we score the support graph based on the weight of the active nodes and edges. Each edge $e(t, h)$ is weighted based on a word-overlap score.\footnote{$w(t, h) = |tok(t) \cap tok(h)| / |tok(h)|$} While \tableilp\ used WordNet~\cite{miller1995wordnet} paths to compute the weight, this measure results in unreliable scores when faced with longer phrases found in Open IE tuples. 

Compared to a curated KB, it is easy to find Open IE tuples that match irrelevant parts of the questions. To mitigate this issue, we improve the scoring of qterms in our ILP objective to focus on important terms. Since the later terms in a question tend to provide the most critical information, we scale qterm coefficients based on their position. Also, qterms that appear in almost all of the selected tuples tend not to be discriminative as any tuple would support such a qterm. Hence we scale the coefficients by the inverse frequency of the tokens in the selected tuples.  

\paragraph{Constraints}\mbox{}\\
Since Open IE tuples do not come with schema and join rules, we can define a substantially simpler model compared to \tableilp. This reduces the reasoning capability but also eliminates the reliance on hand-authored join rules and regular expressions used in \tableilp. We discovered (see empirical evaluation) that this simple model can achieve the same score as \tableilp\ on the Regents test (target test set used by \tableilp) and generalizes better to different grade levels.

\begin{table}[t]
\footnotesize
\centering
\begin{tabular}{l}
\hline
 Active field must have $< w_1$ connected edges\\
 Active choice must have $< w_2$ edges \\
 Active qterm must have $< w_3$ edges \\
 Support graph must have $<w_4$ active tuples\\
 \hline 
 Active tuple must have $\geq w_5$ active fields \\
 Active tuple must have an edge to some qterm \\
 Active tuple must have an edge to some choice \\
\hline
Active tuple must have active subject \\
If a tuple predicate aligns to $q$, the subject (object) must \\
 \ \ \ align to a term preceding (following, resp.) $q$ \\
 \hline
\end{tabular}
\caption{High-level ILP constraints; we report results for $\vec{w} = (2, 4, 4, 4, 2)$; the model can be improved with more careful parameter selection}
\label{const}
\end{table}

We define active vertices and edges using ILP constraints: an active edge must connect two active vertices and an active vertex must have at least one active edge.
To avoid positive edge coefficients in the objective function resulting in spurious edges in the support graph, we limit the number of active edges from an active tuple, question choice, tuple fields, and qterms (first group of constraints in Table \ref{const}). Our model is also capable of using multiple tuples to support different parts of the question as illustrated in Figure~\ref{fig:suppGraph}. To avoid spurious tuples that only connect with the question (or choice) or ignore the relation being expressed in the tuple, we add constraints that require each tuple to connect a qterm with an answer choice (second group of constraints in Table \ref{const}).

We also define new constraints based on the Open IE tuple structure. Since an Open IE tuple expresses a fact about the tuple's subject, we require the subject to be active in the support graph. To avoid issues such as \emph{(Planet; orbit; Sun)} matching the sample question in the introduction (``Which object$\ldots$orbits around a planet''), we also add an ordering constraint (third group in Table \ref{const}). 

\forCR{Its worth mentioning that \tupleinf\ only combines parallel evidence i.e. each tuple must connect words in the question to the answer choice. For reliable multi-hop reasoning using OpenIE tuples, we can add inter-tuple connections to the support graph search, controlled by a small number of rules over the OpenIE predicates. Learning such rules for the Science domain is an open problem and potential avenue of future work.} 

\section{Experiments}

Comparing our method with two state-of-the-art systems for 4th and 8th grade science exams, we demonstrate that (a) \tupleinf\ with only automatically extracted tuples significantly outperforms \tableilp\ with its original curated knowledge as well as with additional tuples, and (b) \tupleinf's complementary approach to \ir\ leads to an improved ensemble. 
Numbers in bold indicate statistical significance based on the Binomial exact test~\cite{howell2012statistical} at $p=0.05$.

We consider two question sets. 
(1) \textbf{4th Grade set} (1220 train, 1304 test) is a 10x larger superset of the NY Regents questions~\cite{aristo2016:combining},  and includes professionally written licensed questions. 
(2) \textbf{8th Grade set} (293 train, 282 test) contains 8th grade questions from various states.\footnote{http://allenai.org/data/science-exam-questions.html}

We consider two knowledge sources. The \textbf{Sentence corpus} (S) consists of domain-targeted $~$80K sentences 
 and 280 GB of plain text extracted from web pages used by \namecite{aristo2016:combining}. This corpus is used by the \ir\ solver and also used to create the tuple KB T and on-the-fly tuples $T'_{qa}$. 
 Additionally, \tableilp\ uses $\sim$70 \textbf{Curated tables} (C) designed for 4th grade NY Regents exams.
%

We compare \tupleinf\ with two state-of-the-art baselines.
\textbf{IR} is a simple yet powerful information-retrieval baseline \cite{aristo2016:combining} that selects the answer option with the best matching sentence in a corpus. 
\textbf{\tableilp} is the state-of-the-art structured inference baseline \cite{tableilp2016} developed for science questions.
\begin{table}
\setlength{\tabcolsep}{10pt}
\setlength{\doublerulesep}{\arrayrulewidth}
\centering
\small
\begin{tabular}{lcc}
Solvers & 4th Grade & 8th Grade \\
\hline \hline
\T \tableilp(C) & 39.9 & 34.1 \\ 
\B \tupleinf(T+T') & \textbf{51.7} & \textbf{51.6} \\
\hline
\T \tableilp(C+T) & 42.1 & 37.9 \\ 
\tupleinf(C+T) & \textbf{47.5} & \textbf{48.0} \\
\end{tabular}
\caption{\tupleinf\ is significantly better at structured reasoning than \tableilp}
\label{struct}
\end{table}

\begin{table}
\setlength{\tabcolsep}{6pt}
\setlength{\doublerulesep}{\arrayrulewidth}
\centering
\small
\begin{tabular}{lcc}
Solvers & 4th Grade & 8th Grade \\
\hline \hline
\T \ir(S) & 52.0 & 52.8 \\ 
\ir(S) + \tableilp(C) & 53.3 & 54.5 \\
\ir(S) + \tupleinf(T+T') & \textbf{55.3} &  55.1 \\
\end{tabular}
\caption{\tupleinf\ is complementarity to \ir, resulting in a strong ensemble}
\label{comp}
\end{table}

\subsection{Results}

Table~\ref{struct} shows that \tupleinf, with no curated knowledge, outperforms \tableilp\ on both question sets by more than 11\%. The lower half of the table shows that even when both solvers are given the same knowledge (C+T),\footnote{See Appendix \ref{app_exp} for how tables (and tuples) are used by \tupleinf\ (and \tableilp).} 
 the improved selection and simplified model of \tupleinf\footnote{On average, \tableilp\ (\tupleinf) has 3,403 (1,628, resp.) constraints and 982 (588, resp.) variables. \tupleinf's ILP can be solved in half the time taken by \tableilp, \forCR{resulting in 68.6\% reduction in overall question answering time.}} results in a statistically significant improvement. 
 Our simple model, \tupleinf(C + T), also achieves scores comparable to \tableilp\ on the latter's target Regents questions (61.4\% vs \tableilp's reported 61.5\%) without any specialized rules. 

Table~\ref{comp} shows that while \tupleinf\ achieves similar scores as the \ir\ solver, the approaches are complementary (structured lossy knowledge reasoning vs.\ lossless sentence retrieval). The two solvers, in fact, differ on 47.3\% of the training questions. To exploit this complementarity, we train an ensemble system~\cite{aristo2016:combining} which, as shown in the table, provides a substantial boost over the individual solvers. Further, \ir\ + \tupleinf\ is consistently better than \ir\ + \tableilp. Finally, in combination with \ir\ and the statistical association based \pmi\ solver \forCR{(that scores 54.1\% by itself)} of \namecite{aristo2016:combining}, \tupleinf\ achieves a score of 58.2\% as compared to \tableilp's ensemble score of 56.7\% on the 4th grade set
, again attesting to \tupleinf's strength.

\section{\forCR{Error Analysis}}
\label{app_err}
We describe four classes of failures that we observed, and the future work they suggest.

\textbf{Missing Important Words}: \textit{Which material will spread out to completely fill a larger container? (A)air (B)ice (C)sand (D)water}\\
In this question, we have tuples that support water will spread out and fill a larger container but miss the critical word ``completely''. An approach capable of detecting salient question words could help avoid that. 

\textbf{Lossy IE}: \textit{Which action is the best method to separate a mixture of salt and water? ...}\\
The IR solver correctly answers this question by using the sentence: \textit{Separate the salt and water mixture by evaporating the water.} However, \tupleinf\ is not able to answer this question as Open IE is unable to extract tuples from this imperative sentence. While the additional structure from Open IE is useful for more robust matching, converting sentences to Open IE tuples may lose important bits of information.

\textbf{Bad Alignment}: \textit{Which of the following gases is necessary for humans \underline{to breathe} in order to live?(A) Oxygen(B) Carbon dioxide(C) Helium(D) Water vapor}\\
\tupleinf\ returns ``Carbon dioxide'' as the answer because of the tuple \textit{(humans; \underline{breathe out}; carbon dioxide)}. The chunk ``to breathe'' in the question has a high alignment score to the ``breathe out'' relation in the tuple even though they have completely different meanings.  Improving the phrase alignment can mitigate this issue. 

\textbf{Out of scope}: \textit{Deer live in forest for shelter. If the forest was cut down, which situation would most likely happen?...}\\
Such questions that require modeling a state presented in the question and reasoning over the state are out of scope of our solver.

\section{Conclusion}
We presented a new QA system, \tupleinf, that can reason over a large, potentially noisy tuple KB to answer complex questions. Our results show that \tupleinf\ is a new state-of-the-art structured solver for elementary-level science that does not rely on curated knowledge and generalizes to higher grades. Errors due to lossy IE and misalignments suggest future work in incorporating context and distributional measures.   

\bibliography{citation}

\begin{thebibliography}{}
\expandafter\ifx\csname natexlab\endcsname\relax\def\natexlab#1{#1}\fi

\bibitem[{Achterberg(2009)}]{solver:scip}
Tobias Achterberg. 2009.
\newblock {SCIP:} solving constraint integer programs.
\newblock {\em Math. Prog. Computation\/} 1(1):1--41.

\bibitem[{Banko et~al.(2007)Banko, Cafarella, Soderland, Broadhead, and
  Etzioni}]{Banko2007OpenIE}
Michele Banko, Michael~J. Cafarella, Stephen Soderland, Matthew Broadhead, and
  Oren Etzioni. 2007.
\newblock Open information extraction from the web.
\newblock In {\em IJCAI\/}.

\bibitem[{Berant et~al.(2013)Berant, Chou, Frostig, and Liang}]{sempre}
J.~Berant, A.~Chou, R.~Frostig, and P.~Liang. 2013.
\newblock Semantic parsing on {F}reebase from question-answer pairs.
\newblock In {\em Empirical Methods in Natural Language Processing (EMNLP)\/}.

\bibitem[{Berant and Liang(2014)}]{parasempre}
Jonathan Berant and Percy Liang. 2014.
\newblock Semantic parsing via paraphrasing.
\newblock In {\em ACL\/}.

\bibitem[{Brill et~al.(2002)Brill, Dumais, and Banko}]{brill2002analysis}
Eric Brill, Susan Dumais, and Michele Banko. 2002.
\newblock An analysis of the {AskMSR} question-answering system.
\newblock In {\em Proceedings of EMNLP\/}. pages 257--264.

\bibitem[{Cheng et~al.(2016)Cheng, Zhu, Wang, Chen, and Qu}]{gaokaoIJCAI}
Gong Cheng, Weixi Zhu, Ziwei Wang, Jianghui Chen, and Yuzhong Qu. 2016.
\newblock Taking up the gaokao challenge: An information retrieval approach.
\newblock In {\em IJCAI\/}.

\bibitem[{Clark(2015)}]{aristo2015:challenge}
Peter Clark. 2015.
\newblock Elementary school science and math tests as a driver for {AI:} take
  the {A}risto challenge!
\newblock In {\em 29th AAAI/IAAI\/}. Austin, TX, pages 4019--4021.

\bibitem[{Clark et~al.(2016)Clark, Etzioni, Khot, Sabharwal, Tafjord, Turney,
  and Khashabi}]{aristo2016:combining}
Peter Clark, Oren Etzioni, Tushar Khot, Ashish Sabharwal, Oyvind Tafjord, Peter
  Turney, and Daniel Khashabi. 2016.
\newblock Combining retrieval, statistics, and inference to answer elementary
  science questions.
\newblock In {\em 30th AAAI\/}.

\bibitem[{Dagan et~al.(2010)Dagan, Dolan, Magnini, and
  Roth}]{dagan2010recognizing}
Ido Dagan, Bill Dolan, Bernardo Magnini, and Dan Roth. 2010.
\newblock Recognizing textual entailment: Rational, evaluation and
  approaches--erratum.
\newblock {\em Natural Language Engineering\/} 16(01):105--105.

\bibitem[{Fader et~al.(2013)Fader, Zettlemoyer, and Etzioni}]{paralex}
Anthony Fader, Luke~S. Zettlemoyer, and Oren Etzioni. 2013.
\newblock Paraphrase-driven learning for open question answering.
\newblock In {\em ACL\/}.

\bibitem[{Fader et~al.(2014)Fader, Zettlemoyer, and Etzioni}]{oqa}
Anthony Fader, Luke~S. Zettlemoyer, and Oren Etzioni. 2014.
\newblock Open question answering over curated and extracted knowledge bases.
\newblock In {\em KDD\/}.

\bibitem[{Ferrucci et~al.(2010)Ferrucci, Brown, Chu-Carroll, Fan, Gondek,
  Kalyanpur, Lally, Murdock, Nyberg, Prager et~al.}]{watson}
David Ferrucci, Eric Brown, Jennifer Chu-Carroll, James Fan, David Gondek,
  Aditya~A Kalyanpur, Adam Lally, J~William Murdock, Eric Nyberg, John Prager,
  et~al. 2010.
\newblock Building {W}atson: An overview of the {DeepQA} project.
\newblock {\em AI Magazine\/} 31(3):59--79.

\bibitem[{Howell(2012)}]{howell2012statistical}
David Howell. 2012.
\newblock {\em Statistical methods for psychology\/}.
\newblock Cengage Learning.

\bibitem[{Khashabi et~al.(2016)Khashabi, Khot, Sabharwal, Clark, Etzioni, and
  Roth}]{tableilp2016}
Daniel Khashabi, Tushar Khot, Ashish Sabharwal, Peter Clark, Oren Etzioni, and
  Dan Roth. 2016.
\newblock Question answering via integer programming over semi-structured
  knowledge.
\newblock In {\em IJCAI\/}.

\bibitem[{Khot et~al.(2015)Khot, Balasubramanian, Gribkoff, Sabharwal, Clark,
  and Etzioni}]{aristo2015:mln}
Tushar Khot, Niranjan Balasubramanian, Eric Gribkoff, Ashish Sabharwal, Peter
  Clark, and Oren Etzioni. 2015.
\newblock Exploring {M}arkov logic networks for question answering.
\newblock In {\em EMNLP\/}.

\bibitem[{Kwiatkowski et~al.(2013)Kwiatkowski, Choi, Artzi, and
  Zettlemoyer}]{kwiatkowski13}
Tom Kwiatkowski, Eunsol Choi, Yoav Artzi, and Luke~S. Zettlemoyer. 2013.
\newblock Scaling semantic parsers with on-the-fly ontology matching.
\newblock In {\em EMNLP\/}.

\bibitem[{Miller(1995)}]{miller1995wordnet}
George Miller. 1995.
\newblock Wordnet: a lexical database for english.
\newblock {\em Communications of the ACM\/} 38(11):39--41.

\bibitem[{Pasupat and Liang(2015)}]{tableParse}
Panupong Pasupat and Percy Liang. 2015.
\newblock Compositional semantic parsing on semi-structured tables.
\newblock In {\em ACL\/}.

\bibitem[{Richardson and Domingos(2006)}]{richardson2006markov}
Matthew Richardson and Pedro Domingos. 2006.
\newblock {M}arkov logic networks.
\newblock {\em Machine learning\/} 62(1--2):107--136.

\bibitem[{Sharma et~al.(2015)Sharma, Vo, Aditya, and Baral}]{knowlhunting}
Arpit Sharma, Nguyen~Ha Vo, Somak Aditya, and Chitta Baral. 2015.
\newblock Towards addressing the winograd schema challenge - building and using
  a semantic parser and a knowledge hunting module.
\newblock In {\em IJCAI\/}.

\bibitem[{Sun et~al.(2016)Sun, Ma, He, tau Yih, Su, and Yan}]{tabCell}
Huan Sun, Hao Ma, Xiaodong He, Wen tau Yih, Yu~Su, and Xifeng Yan. 2016.
\newblock Table cell search for question answering.
\newblock In {\em WWW\/}.

\bibitem[{Yin et~al.(2015)Yin, Duan, Kao, Bao, and Zhou}]{taqa}
Pengcheng Yin, Nan Duan, Ben Kao, Jun-Wei Bao, and Ming Zhou. 2015.
\newblock Answering questions with complex semantic constraints on open
  knowledge bases.
\newblock In {\em CIKM\/}.

\end{thebibliography}
\bibliographystyle{acl_natbib}

\clearpage
\appendix
\section{Appendix: ILP Model Details}
\label{app_model}
To build the ILP model, we first need to get the questions terms (qterm) from the question by chunking the question using an in-house chunker based on the postagger from FACTORIE.~\footnote{http://factorie.cs.umass.edu/}
\paragraph{Variables}\mbox{}\\
The ILP model has an active vertex variable for each qterm ($x_q$), tuple ($x_t$), tuple field ($x_f$) and question choice ($x_a$). Table \ref{coeff} describes the coefficients of these active variables. For example, the coefficient of each qterm is a constant value (0.8) scaled by three boosts. The idf boost, $idfB$ for a qterm, x is calculated as $\log(1 + (|T_{qa}| + |T'_{qa}|)/n_x)$ where $n_x$ is the number of tuples in $T_{qa} \cup T'_{qa}$ containing x. The science term boost, $scienceB$ boosts coefficients of qterms that are valid science terms based on a list of ~9K terms. The location boost, $locB$ of a qterm at index $i$ in the question is given by $i/tok(q)$ (where $i$=1 for the first term).

Similarly each edge, $e$ has an associated active edge variable with the word overlap score as its coefficient, $c_e$. For efficiency, we only create qterm$\rightarrow$field edge and field$\rightarrow$choice edge if the coefficient is greater than a certain threshold (0.1 and 0.2, respectively). Finally the objective function of our ILP model can be written as:\\
$$
\sum_{q \in \mathrm{qterms}} c_q x_q
  + \sum_{t \in \mathrm{tuples}} c_t x_t
  + \sum_{e \in \mathrm{edges}} c_e x_e
$$
\begin{table}[h!t]
\setlength{\tabcolsep}{5pt}
\setlength{\doublerulesep}{\arrayrulewidth}
\centering
\begin{tabular}{lll}
Description & Var. & Coefficient (c) \\
\hline \hline
\T Qterm & $x_q$ & 0.8$\cdot$idfB$\cdot$scienceB$\cdot$locB \\
Tuple & $x_t$ & -1 + jaccardScore(t, qa) \\
Tuple Field & $x_f$ & 0 \\
Choice & $x_a$ & 0
\end{tabular}
\caption{Coefficients for active variables. }
\label{coeff}
\end{table}

\paragraph{Constraints}
Next we describe the constraints in our model. We have basic definitional constraints over the active variables.
\begin{table}[ht]
\begin{tabular}{l}
\hline
Active variable must have an active edge\\
Active edge must have an active source node\\
Active edge must have an active target node\\
Exactly one answer choice must be active\\
Active field implies tuple must be active \\
\hline
\end{tabular}
\end{table}

Apart from the constraints described in Table \ref{const}, we also use the \emph{which-term} boosting constraints defined by \tableilp\ (Eqns. 44 and 45 in Table 13~\cite{tableilp2016}). As described in Section \ref{app_exp},  we create a tuple  from table rows by setting pairs of cells as the subject and object of a tuple. For these tuples, apart from requiring the subject to be active, we also require the object of the tuple. This would be equivalent to requiring at least two cells of a table row to be active.

\section{Experiment Details}
\label{app_exp}
We use the SCIP ILP optimization engine~\cite{solver:scip} to optimize our ILP model. To get the score for each answer choice $a_i$, we force the active variable for that choice $x_{a_i}$ to be one and use the objective function value of the ILP model as the score. For evaluations, we use a 2-core 2.5 GHz Amazon EC2 linux machine with 16 GB RAM. To evaluate \tableilp\ and \tupleinf\ on curated tables and tuples, we converted them into the expected format of each solver as follows. 
\subsection{Using curated tables with \tupleinf}
For each question, we select the 7 best matching tables using the tf-idf score of the table w.r.t. the question tokens and top 20 rows from each table using the Jaccard similarity of the row with the question. (same as ~\namecite{tableilp2016}). We then convert the table rows into the tuple structure using the relations defined by \tableilp. For every pair of cells connected by a relation, we create a tuple with the two cells as the subject and primary object with the relation as the predicate. The other cells of the table are used as additional objects to provide context to the solver. We pick top-scoring 50 tuples using the Jaccard score.

\subsection{Using Open IE tuples with \tableilp}
We create an additional table in \tableilp\ with all the tuples in $T$. Since \tableilp\ uses fixed-length $(subject; predicate; object)$ triples, we need to map tuples with multiple objects to this format. For each object, $O_i$ in the input Open IE tuple $(S; P; O_1; O_2 \ldots)$, we add a triple $(S; P; O_i)$ to this table.

\end{document}